# Deep Learning Models in Detection of Dietary Supplement Adverse Event Signals from Twitter

Yefeng Wang[1], Yunpeng Zhao[2], Jiang Bian[2], Rui Zhang[1,3]*

[1]Institute for Health Informatics, University of Minnesota, Minneapolis, MN, USA, [2]Department of Health Outcomes & Biomedical Informatics, University of Florida, Gainesville, FL, USA, [3]Department of Pharmaceutical Care & Health Systems, University of Minnesota, Minneapolis, MN, USA

**Corresponding author:**

*Rui Zhang, PhD

Institute for Health Informatics and Department of Pharmaceutical Care & Health Systems, University of Minnesota

516 Delaware St SE, Minneapolis, MN 5545

Email: zhan1386@umn.edu

Office Phone: 612-626-4209

**Abstract**

**Objective**: The objective of this study is to develop a deep learning pipeline to detect signals on dietary supplement-related adverse events (DS AEs) from Twitter.

**Material and Methods**: We obtained 247,807 tweets ranging from 2012 to 2018 that mentioned both DS and AE. We annotated biomedical entities and relations on 2,000 randomly selected tweets. For the concept extraction task, we compared the performance of traditional word embeddings with SVM, CRF and LSTM-CRF classifiers to BERT models. For the relation extraction task, we compared GloVe vectors with CNN classifiers to BERT models. We chose the best performing models in each task to assemble an end-to-end deep learning pipeline to detect DS AE signals and compared the results to the known DS AEs from a DS knowledge base (i.e., iDISK).

**Results**: In both tasks, the BERT-based models outperformed traditional word embeddings. The best performing concept extraction model is the BioBERT model that can identify supplement, symptom, and body organ entities with F1-scores of 0.8646, 0.8497, and 0.7104, respectively. The best performing relation extraction model is the BERT model that can identify purpose and AE relations with F1-scores of 0.8335 and 0.7538, respectively. The end-to-end pipeline was able to extract DS indication and DS AEs with an F1-score of 0.7459 and 0,7414, respectively. Comparing to the iDISK, we could find both known and novel DS-AEs.

**Conclusion**: We have demonstrated the feasibility of detecting DS AE signals from Twitter with a BioBERT-based deep learning pipeline.

# INTRODUCTION

Dietary supplements (DSs) are gaining popularity depicted by their steady escalating usage which reaches all-time high in 2019 according to an annual survey on consumers' DS usage conducted by the Council for Responsible Nutrition (CRN). Seventy-seven percent of Americans have used at least one DS, and adults between the ages 35 – 54 have the highest usage of DS. [1] However, DS regulatory policies are different and less rigorous than those covering their drug counterpart. As per the Dietary Supplement Health and Education Act of 1994 (DSHEA), [2] both DS products and DS ingredients are regulated by the Food and Drug Administration (FDA), but clinical approval trials for DS safety and efficacy are not mandatory. [3] As a result, there is an estimated over 23,000 emergency department visits per year were attributed to DS use. [4]

The existing pharmacovigilance infrastructure around DS primarily relies on post-marketing spontaneous reporting system (SRS) where DS manufacturers, researchers, clinicians, and consumers voluntarily report AEs online. In the US, the Center for Food Safety and Applied Nutrition (CFSAN) under the FDA launched the CFSAN Adverse Event Reporting System (CAERS) in 2003 to facilitate post-market monitoring and surveillance of adverse event reports (AERs) associated with food, cosmetic, and DS. [5] The primary purpose of CAERS is to enhance consumer safety through real-time assessment of AERs. [6] However, the distribution of the AER reporting sources is heavily skewed. [7] By Q2 of 2016, healthcare professionals had contributed 237,996 AERs, while DS consumers only contributed only 69,267 AERs to the SRS. [8] It is possible that the consumers may not know how to use the SRS to report the AEs they experienced, or they might not be aware that such systems exist. Moreover, when health professionals report adverse events (AEs), they focus more on serious AEs that may present grave danger to patient health but overlook more common AEs that are not life-threatening. [9]

Therefore, additional data sources that put more emphasis on DS consumers are necessary for effective DS surveillance and monitoring.

Social media (SM) has emerged as a valuable resource for pharmacovigilance. [10, 11] As a series of web-based platforms that facilitate easy and instant communication between individuals, social media has made real-time self-reports on health-related behaviors possible, including DS use. A recent paper has proposed to use SM data for pharmacovigilance, for its adequate representation of mild symptom DS AEs making it an ideal complement to balance reporting bias in SRSs. [12] SM may also include other important health information and health behaviors that are often not available through SRSs, e.g., mentions of smoking status, [13] allergies, [14] pregnancy, [15] and/or presence of other comorbidities. [16, 17]

The identification of DS AEs from text data is a two-step process. Firstly, the terms that correspond to DSs and AEs are determined, which is a concept extraction task; and then, the relations between these terms are examined to see if the AE is directly caused by DS usage, which is a relation extraction task. Various corpora have been developed to train machine learning models and benchmark their performances on both tasks, such as the ones from Integrating Biology and the Bedside (i2b2), [18] ShARe/CLEF, [19] and SemEval challenges. [20] The results have shown that the vector representation of the words, or word embeddings, played an important role in achieving state-of-the-art performance. [21] Traditional word embeddings such as Word2Vec, [22] GloVe, [23] and fastText [24] managed to integrate word context information into a single vector. This presents an obstacle to the concept extraction task and the relation extraction task as the single vector cannot cope with polysemy. For example, the word "*cut*" can either be a noun meaning "*a flesh wound caused by a sharp object*", or be a verb referring to the action of "*penetrate with an edged instrument*", or be a part of a verbal phrase such as "*cut back*" which means "*to reduce*". However, Word2Vec, GloVe, and fastText could only learn a final word vector for the word "cut" regardless of its meaning in the text. It would be hard to isolate the "*flesh wound*" meaning of the word from the other ones if the same word vector were used for all future predictions. Contextual embeddings such as ELMo (Embeddings from Language Models) [25] and BERT (Bidirectional Encoder Representations from Transformers) [26] that can alter dynamically based on its meaning in context were designed to overcome such drawback. Instead of a final real-valued vector, contextual embeddings use all layers of a deep neural network trained on a large corpus. In this study, we focused on BERT as it employs a fine-tuning approach, where the pre-trained deep neural network is further optimized with respect to the downstream tasks. BERT models have shown superior performance than traditional word embeddings in natural language processing (NLP) tasks such as concept extraction, relation extraction, and question answering tasks. [27] However, these models were trained on general domain texts such as Wikipedia, which may lead to underperformance on biomedical NLP tasks. An adaptation of BERT model, BioBERT, was designed to improve on this limitation by adding PubMed abstracts and PMC full-text articles to the training corpus. [28] BERT and BioBERT models have been applied to social media for identification of personal health experience [29] and drug AE mining, [30] but to the best of our knowledge, these state-of-the-art word embeddings have not been applied in detecting DS AE signals from Twitter.

Thus, in this study, our main contributions are: (1) to evaluate traditional and contextual word embeddings and deep learning models on the annotated twitter data, (2) to assess the feasibility of deep learning models on Twitter to detect DS-AEs.

**MATERIAL AND METHODS**

The overview of the methods was shown in Figure 1. We collected and annotated a set of DS-related tweets and compared the performance of traditional embeddings with contextual embeddings in the concept extraction and relation extraction tasks, respectively. The best performing models were used to assemble an end-to-end pipeline for the identification of DS AEs from tweets. We compared the signals generated by the pipeline on a larger corpus with an existing DS knowledge base (i.e., integrated dietary supplement knowledge base (iDISK)). [31]

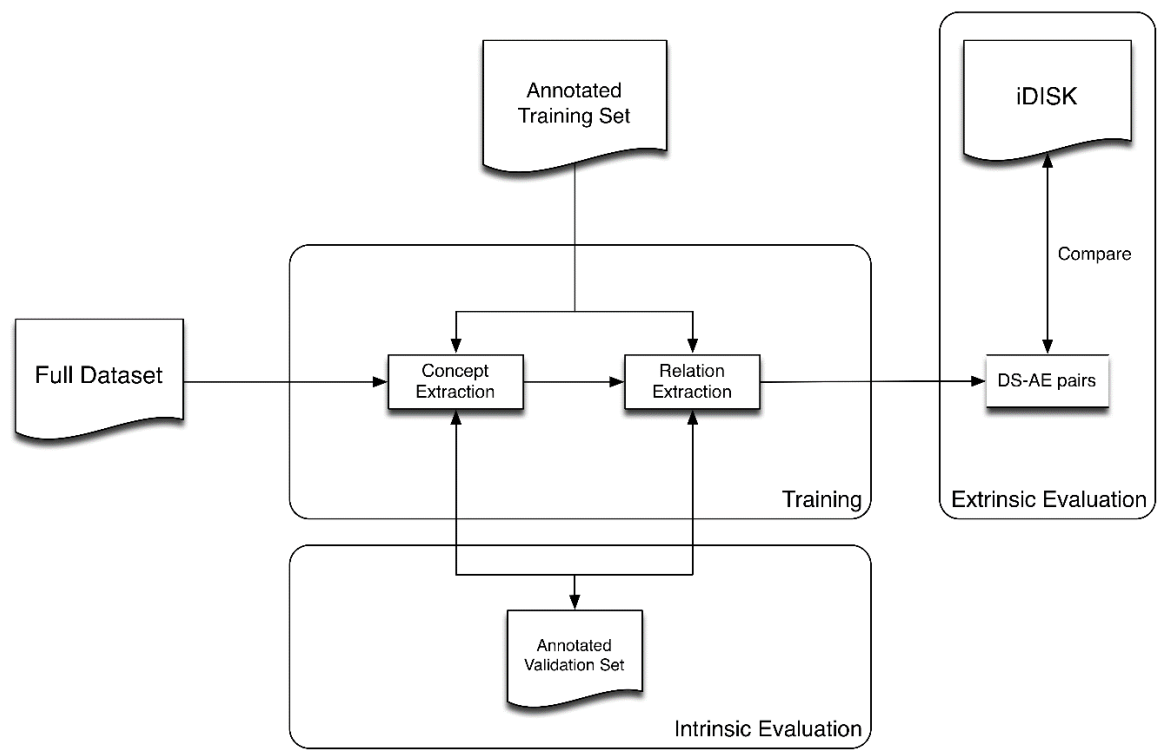

Figure 1. The overview of our study workflow.

**Data Collection**

To retrieve the tweets with co-occurrence of DS and symptom/body organ terms, we compiled two lists of terms. The DS term list was obtained from our previous study, which contains 332 DS terms including 31 commonly used DS names and their name variants [32]. The symptom/body organ term list contains 14,143 terms by integrating the ADR lexicon [33] and the iDISK knowledge base. We selected only English tweets to develop the concept extraction and relation extraction models. A total of 247,807 tweets that satisfy the criteria were found from a Twitter database we constructed in prior work [34, 35] using the Twitter streaming application programming interface (API), covering daily publictweets from 2012 to 2018.

**Data preprocessing**

We employed the *ekphrasis* package [36] to remove uniform resource locators (URLs), user handle (e.g. @username), hashtag symbol ("#"), and emoji characters. Contractions such as "*doesn't*", "*won't*" were expanded into "*does not*" and "*will not*", respectively. Hashtags were segmented into their constituent words (e.g. "*ILoveVitaminC*" would be segmented into "I *Love Vitamin C*"). Stop words were not removed, because some of the stop words are meaningful (e.g., "*throw up*" is a common phrase to describe vomiting, but after removing the stop words it would become "*throw*", losing the original meaning). The tweets were then lower-cased, and we used *SpaCy* package [37] to derive the part-of-speech tags of every word in the tweet.

**Annotation**

We randomly selected 2,000 tweets from the dataset for annotation. Two annotators manually reviewed the tweets, highlighting all biomedical entities and relations between them with brat annotation tool. Initially, a random sample of 100 tweets were selected for creation of annotation guideline and calculation of inter-rater agreement. We describe the detailed annotation guideline in the following subsection.

*Concept Extraction:* The Beginning-Inside-Outside (BIO) representation was selected to label the entities. A concrete annotation example is given in Figure 2. Note that *O* tags were not highlighted in this example.

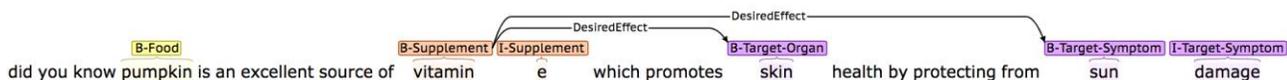

Figure 2. An annotation example.

We defined three entity types: DS, symptoms, and body organs. Supplements include both oral DS and supplements that are topically applied. For example, Vitamin E taken orally, and Vitamin E oil used on skin should all be included. The form of supplement is kept as a part of the named entity. For example, the word "*oil*" in "*oregano oil*" should not be omitted. For herbal supplement, it is common that the supplement is a part of the plant. The word that describes the part is also kept within the named entity. For example, the word "*seed*" needs to be annotated in the noun phrase "*grape seed*". We annotated the deficiency of the supplement, e.g., "*vitamin b12 deficiency increases risk of cancer*", as "*deficiency*". With the deficiency information available, the machine learning model can avoid making the mistake of identifying cancer as an AE of vitamin B12 in this example.

Symptoms are entities that describe the specifics of a DS AE, such as "*cold*", "*cough*", "*diarrhea*", "*cancer*", "*throw up*", "*feel sick*", etc. However, some tweets did not specify the symptoms of a DS AE. For example, in the tweet "*too much vitamin A is bad for your liver*", although no symptoms were mentioned, the body organs where the supplement might take effect could indicate a DS AE Therefore, we would annotate body organ entities as a DS AE signal as well.

*Relation Extraction:* We defined two binary relations based on the above definition of entities: indication and adverse events. Purposes are positive effects due to the use of the supplements. For example, "*Vitamin D reduces fatigue*" implies that the purpose of using Vitamin D is to deal with fatigue. "*Kava kava balances mood*" indicates that the purpose of using "*kava kava*" is to balance mood. AEs are negative effects due to the use of the supplements. In the example of "*Excess vitamin D weakens bone*", Vitamin D has an undesired effect on bone. Iron has resulted an AE – queasy in the "*Iron pills seems to make me feel queasy*".

**Concept Extraction Models**

In this task, we compared machine learning and deep learning models to recognize biomedical entities including DS names, AE terms, and body organs. Among 2,000 tweets, 70% of which was used to training the model, 10% for parameter optimization, and the remaining 20% for evaluation.

The performance of concept extraction is dependent on the word representation. We compared four types of traditional word embeddings, i.e., PubMed Word2Vec (trained on 5.5 billion tokens including English Wikipedia, PubMed abstracts, and PMC full-text), [38] FastText (trained on 600 billion tokens), [24] GloVe Twitter vectors (trained on 27 billion tokens), [23] and GloVe Crawl word vectors (trained on 840 billion tokens) [23], with two types of contextual word embeddings, BERT and BioBERT.

When using traditional word embeddings for the concept extraction task, we have compared three models, including support vector machine (SVM), conditional random field (CRF), and long-short term memory with a CRF layer (LSTM-CRF). For SVM classifier, we used linear kernel with $L_2$-regularization without further optimization as a baseline; for CRF classifier, we used L-BFGS optimization with Elastic Net regularization to train the model; for LSTM-CRF, we used a 64-dimension bidirectional LSTM layer to encode the word embeddings and used Viterbi algorithm for CRF decoding. No dropout was applied to the LSTM layer. We used Adam optimization to train the model with a learning rate of 0.001 and a weight decay of $10^{-4}$. When using BERT and BioBERT embeddings, a softmax classifier layer was appended to the BERT layers and then trained on the annotated dataset for fine-tuning. Adam optimization with a learning rate of $5\times10^{-5}$ was used to train the model. For LSTM-CRF, we trained the model for 40 epochs with batch size of 32; for BERT and BioBERT models, we trained the models for 10 epochs.

We measured precision, recall, and F1-score based on the metrics defined by the International Workshop on Semantic Evaluation (SemEval). [39] The evaluation metrics are shown in Table 1.

Table 1. Possible scenario of model prediction compared to the gold standard. MIS = mismatch, SPU = spurious, COR = correct, PAR = partial, INC = incorrect.

| Gold Standard | | Model Prediction | | Evaluation |
|---|---|---|---|---|
| Term | Entity Type | Term | Entity Type | |
| flaxseed | Supplement | | | MIS |

|  |  | headache | Symptom | SPU |
|---|---|---|---|---|
| vitamin C | Supplement | vitamin | Supplement | PAR |
| folate | Supplement | folate | Supplement | COR |
| headache | Symptom | headache | Supplement | INC |

Based on the notations in Table 1, the precision of a certain entity type can be calculated as:

$$Precision = \frac{COR + 0.5 \times PAR}{COR + INC + PAR + SPU}$$

The recall can be calculated as:

$$Recall = \frac{COR + 0.5 \times PAR}{COR + INC + PAR + MIS}$$

F1-score can thus be calculated as:

$$F1 = \frac{2 \times Precision \times Recall}{Precision + Recall}$$

The above calculations were performed for each entity type.

To compare the statistical significance of the performance of LSTM-CRF, BERT, and BioBERT models, we trained models 20 times and calculated the mean and the standard deviation of precision, recall, and F1-score. We performed paired $t$-tests on F1-scores to test the statistical significance of the model performance difference, where a $p$-value < 0.001 was considered statistically significant.

**Relation Extraction Models**

The concept extraction model helped find the position of biomedical entities within a tweet. The next step is to identify the relations between the biomedical entities, especially between the DS and AE entities. Multiple relations exist between a DS and an AE terms. In the tweet "*sounds weird but it works because vitamin C helps with sore throat*", the relation between the DS "*vitamin C*" term and the AE term "*sore throat*" is an indication, i.e., the supplement was used with the intention to treat a symptom. But in the tweet "*note to self if you are used to 250 mg of niacin jump up to 500 mg the niacin flush is so intense*", the relation between "*niacin*" and "*flush*" is an adverse event, i.e. the supplement caused the symptom. The relation extraction model should be able to differentiate DS AEs from DS indications. To measure the performance of relation extraction model alone, we developed and compared models which predict relations (i.e., "no relation", "indication", or "AE") between all possible pairs between DS and AE entities.

We used two 50-dimension GloVe vectors as the traditional word embeddings with a single-layer CNN (with pooling) baseline. The word embeddings were trained on Twitter (27 billion tokens) and Wikipedia + GigaWord (6 billion tokens), respectively. As for contextual word embeddings, we still used BERT and BioBERT models and the relation classification was performed by appending a softmax classifier.

We used the *OpenNRE* package to implement the relation extraction model. [40] For CNN classifier, we used a 256-dimension convolutional layer with a kernel size of 3 followed by a max-pooling layer. The dropout applied to the convolutional layer was 0.2. The model was trained with an Adam optimizer with a learning rate of $10^{-4}$ and a weight decay of $10^{-5}$. For BERT and BioBERT models, a softmax classifier layer was appended to the BERT layers and then trained on the annotated dataset for fine-tuning. Adam optimization with a learning rate of

$3 \times 10^{-5}$ was used to train the model. We trained 40 epochs for the CNN model and 25 epochs for the BERT and BioBERT models.

We trained the model 20 times on the annotated tweet dataset with differently randomized training and test set split and tested the statistical significance of the model performance difference We calculated the mean and the standard deviation of precision, recall, and F1-score and performed paired *t*-test on F1-score.

**The End-to-End Model**

We integrated the best performing models for concept extraction and relation extraction for end-to-end DS AEs signal detection. The performance of the pipeline was evaluated by correct predictions of relations.

**Evaluation of Extracted DS AEs Against the iDISK**

We applied the end-to-end model trained on the annotated tweets to the full dataset and compared the signal detected from the machine learning pipeline to DS adverse events recorded in iDISK.

We tallied the occurrences of DS AE pairs extracted by the end-to-end pipeline and selected 200 most-frequently mentioned DS AEs in tweets. We compared these 200 DS AEs to the ones in iDISK, which include symptoms and body organs. For every DS AE pair, we also selected several example tweets to check if the pair was correctly referring to a DS AE.

**RESULTS**

**Data Statistics**

The 2,000 tweets in the entire training set contained 2,244 DS entities, 2,003 symptom entities, and 287 body organ entities. There are 1,471 indication mentions and 442 AE mentions. The inter-rater agreement (kappa score) for the NER task is 0.9416 and 0.8299 for RE task.

**Concept Extraction Task**

Table 2 shows the performance comparison among all the methods and word embeddings used in this task. SVM using traditional word embeddings generally underperformed for all entity types. CRF with the same word embeddings improved the performance, and when coupled with a LSTM network, the LSTM-CRF model using GloVe Crawl embeddings achieved the best F1 scores among those methods using traditional word embeddings for the extraction of supplements, symptoms, and body organ concepts. However, the performance of the extraction of symptoms and body organ concepts using traditional word embeddings was not ideal. BERT and BioBERT models had improved performance in symptoms and body organ concept extraction. The best performing model have a mean F1-score of 0.8526 for symptoms (BERT model) and 0.7104 for body organs (BioBERT) and both F1-scores are significantly better than the LSTM-CRF models with traditional word embeddings ($p < 0.001$). The performance of supplement concept extraction using BERT models was slightly better than BioBERT.

The BERT and BioBERT models have significantly outperformed the LSTM-CRF models with PubMed Word2Vec, fastText, and GloVe Twitter embeddings, but not the GloVe Crawl embeddings. When comparing the BERT and BioBERT models, no statistically significant difference in performance was observed. Therefore, we chose BioBERT model for concept extraction for the end-to-end pipeline due to its higher mean F1-score in body organ concept extraction.

Table 2. Performance of models for extracting supplements, symptoms and body organs concepts.

|  | Supplement | | | Symptoms | | | Organs | | |
| --- | --- | --- | --- | --- | --- | --- | --- | --- | --- |
| Model | Precision | Recall | F1 | Precision | Recall | F1 | Precision | Recall | F1 |
| SVM + PubMed Word2Vec | 0.7652 ± 0.0176 | 0.8338 ± 0.0245 | 0.7979 ± 0.0183 | 0.6265 ± 0.0196 | 0.6409 ± 0.0285 | 0.6334 ± 0.0214 | 0.4388 ± 0.0775 | 0.6710 ± 0.0612 | 0.4915 ± 0.0885 |
| SVM + FastText | 0.7841 ± 0.0182 | 0.8354 ± 0.0218 | 0.7974 ± 0.0231 | 0.6402 ± 0.0241 | 0.6953 ± 0.0176 | 0.6664 ± 0.0186 | 0.4422 ± 0.0563 | 0.4624 ± 0.0715 | 0.4512 ± 0.0614 |
| SVM + GloVe Twitter | 0.7446 ± 0.0211 | 0.7913 ± 0.0285 | 0.7671 ± 0.0226 | 0.6119 ± 0.0167 | 0.6594 ± 0.0157 | 0.6346 ± 0.0141 | 0.4347 ± 0.0639 | 0.4752 ± 0.0921 | 0.4526 ± 0.0745 |
| SVM + GloVe Crawl | 0.7692 ± 0.0215 | 0.8280 ± 0.0269 | 0.7974 ± 0.0231 | 0.6193 ± 0.0170 | 0.6997 ± 0.0181 | 0.6570 ± 0.0160 | 0.4233 ± 0.0654 | 0.5458 ± 0.0860 | 0.4757 ± 0.0708 |
| CRF + PubMed Word2Vec | 0.8759 ± 0.0202 | 0.8197 ± 0.0263 | 0.8466 ± 0.0212 | 0.7755 ± 0.0267 | 0.6959 ± 0.0233 | 0.7334 ± 0.0229 | 0.6598 ± 0.0465 | 0.6013 ± 0.0485 | 0.6272 ± 0.0330 |
| CRF + FastText | 0.8903 ± 0.0217 | 0.7997 ± 0.0263 | 0.8424 ± 0.0226 | 0.8203 ± 0.0217 | 0.6483 ± 0.0223 | 0.7242 ± 0.0212 | 0.6113 ± 0.0655 | 0.3844 ± 0.0625 | 0.4697 ± 0.0607 |
| CRF + GloVe Twitter | 0.8609 ± 0.0170 | 0.7963 ± 0.0214 | 0.8272 ± 0.0170 | 0.7755 ± 0.0267 | 0.6959 ± 0.0233 | 0.7334 ± 0.0229 | 0.6005 ± 0.0477 | 0.5493 ± 0.0678 | 0.5720 ± 0.0508 |
| CRF + GloVe Crawl | 0.8722 ± 0.0182 | 0.8362 ± 0.0208 | 0.8538 ± 0.0186 | 0.8021 ± 0.0184 | 0.7326 ± 0.0244 | 0.7656 ± 0.0193 | 0.6474 ± 0.0477 | 0.6487 ± 0.0580 | 0.6467 ± 0.0403 |
| LSTM-CRF + PubMed Word2Vec | 0.8587 ± 0.0211 | 0.8055 ± 0.0280 | 0.8310 ± 0.0218 | 0.7909 ± 0.0188 | 0.6794 ± 0.0258 | 0.7306 ± 0.0173 | 0.6024 ± 0.0640 | 0.5035 ± 0.0647 | 0.5469 ± 0.0583 |
| LSTM-CRF + FastText | 0.8538 ± 0.0160 | 0.8092 ± 0.0231 | 0.8308 ± 0.0175 | 0.7784 ± 0.0247 | 0.6841 ± 0.0271 | 0.7277 ± 0.0182 | 0.4553 ± 0.0741 | 0.3459 ± 0.0766 | 0.3911 ± 0.0726 |
| LSTM-CRF + GloVe Twitter | 0.8491 ± 0.0321 | 0.8127 ± 0.0196 | 0.8300 ± 0.0179 | 0.8048 ± 0.0150 | 0.6994 ± 0.0244 | 0.7482 ± 0.0155 | 0.6221 ± 0.0612 | 0.5285 ± 0.0763 | 0.5702 ± 0.0636 |
| LSTM-CRF + GloVe Crawl | 0.8736 ± 0.0210 | 0.8375 ± 0.0152 | 0.8551 ± 0.0157 | 0.8012 ± 0.0205 | 0.7146 ± 0.0344 | 0.7550 ± 0.0232 | 0.6298 ± 0.0742 | 0.5383 ± 0.0754 | 0.5793 ± 0.0714 |
| BERT | 0.8560 ± 0.0185 | 0.8736 ± 0.0198[a] | 0.8647 ± 0.0184[a] | 0.8393 ± 0.0161 | 0.8664 ± 0.0147[a] | 0.8526 ± 0.0138[a] | 0.6721 ± 0.0549[a] | 0.7317 ± 0.0568 | 0.6993 ± 0.0481 |
| BioBERT | 0.8570 ± 0.0248[a] | 0.8725 ± 0.0212 | 0.8646 ± 0.0220 | 0.8416 ± 0.0204[a] | 0.8582 ± 0.0200 | 0.8497 ± 0.0172 | 0.6710 ± 0.0612 | 0.7578 ± 0.0562[a] | 0.7104 ± 0.0510[a] |

[a] Best performing model. BERT and BioBERT model outperformed SVM, CRF, and LSTM-CRF models at $p < 0.001$ level. However, the performance of BioBERT is not significantly different from that of BERT model.

**Relation Extraction Task**

Table 3 shows the performance of relation extraction task. While the CNN models were able to achieve comparable performance in DS indication identification, it failed to successfully identify DS AEs and suffered from low recall. BERT models proved to be a better solution with its fine-tuning approach. The best model which used BioBERT contextual embeddings with entity heading features were able to achieve a mean recall of 0.7845 in DS AE extraction, and thus improve the final mean F1-score of 0.7538.

The paired *t*-test shown that the outperformance of BERT models over traditional word embeddings with CNN models was statistically significant. However, among the BERT models, no statistically significant difference was detected. In this case, we chose the BioBERT model with entity headings as our relation extraction model for the end-to-end pipeline as it had higher mean F1-score in both relation categories.

Table 3 Performance of relation extraction models for DS AEs.

| | DS-Indications | | | DS-AEs | | |
|---|---|---|---|---|---|---|
| **Model** | **Precision** | **Recall** | **F1** | **Precision** | **Recall** | **F1** |
| CNN + GloVe Twitter (50d) | 0.7774 ± 0.0252 | 0.7946 ± 0.0318 | 0.7850 ± 0.0124 | 0.6995 ± 0.0653 | 0.6381 ± 0.0539 | 0.6645 ± 0.0410 |
| CNN + GloVe Wikipedia + GigaWord (50d) | 0.7720 ± 0.0206 | 0.7901 ± 0.0280 | 0.7804 ± 0.0142 | 0.7069 ± 0.0553 | 0.5995 ± 0.0783 | 0.6456 ± 0.0561 |
| BERT | 0.8181 ± 0.0319[a] | 0.8522 ± 0.0409 | 0.8335 ± 0.0169 | 0.7312 ± 0.0694 | 0.7845 ± 0.0487[a] | 0.7538 ± 0.0376[a] |
| BioBERT | 0.8177 ± 0.0214 | 0.8595 ± 0.0321[a] | 0.8374 ± 0.0147[a] | 0.7349 ± 0.0430[a] | 0.7603 ± 0.0519 | 0.7459 ± 0.0341 |

[a] Best performing model. BERT and BioBERT model outperformed CNN models at $p < 0.001$ level. However, the performance of BioBERT is not significantly different from that of BERT model.

**End-to-End Pipeline**

We chose BioBERT model for concept extraction and BERT model with entity headings features for relation extraction due to their outperformance over all other models in each task. We assembled them into an end-to-end pipeline, i.e. the output of the concept extraction model will be directly used as inputs into the relation extraction model. It is expected that the error would propagate along the pipeline and thus lead to an overall lower F1-score, and the evaluation results shown in Table 4 confirmed our expectations.

Table 4 Performance of the end-to-end DS AEs extraction pipeline.

| Relation Type | Precision | Recall | F1 |
| --- | --- | --- | --- |
| Indication-DS | 0.7250 ± 0.0498 | 0.7713 ± 0.0452 | 0.7459 ± 0.0321 |
| AE-DS | 0.7217 ± 0.0665 | 0.7644 ± 0.0861 | 0.7414 ± 0.0699 |

**Comparison to iDISK**

We applied the DS AE identification pipeline to the tweets in the entire dataset. The pipeline was able to find 2,159 DS AE pairs from 15,399 source tweets and 13,780 DS indication pairs from 191,642 source tweets. Table 5 presents the 15 most frequently mentioned DS AE pairs in our dataset, while Table 6 presents examples of frequently mentioned DS indication pairs.

Table 5. Examples of most frequently DS effects detected by end-to-end deep learning pipeline.

| DS AE Pairs | Frequency | In iDISK? | Tweet examples |
| --- | --- | --- | --- |
| Vitamin C - sick | 183 | Yes | *"who knew too much vitamin c can make you sick i am upset"* |
| Vitamin B – lung cancer | 172 | No | *"high doses of vitamin b supplements can cause lung cancer in men"* |
| Vitamin C – kidney stones | 158 | Yes | *"some medications yes even prolonged high dose vitamin c causes kidney stones"* |
| Vitamin C - nausea | 129 | Yes | *"too much vitamin c or zinc could cause nausea diarrhea and stomach cramps check your dose"* |
| Vitamin C - diarrhea | 117 | Yes | *"i would eat this whole bag of oranges but vitamin c in high doses can induce skin breakouts and diarrhea facts"* |
| Fish oil – prostate cancer | 116 | No | *"saw the one on fish oil being linked to prostate cancer helps keep cholesterol down but gives you cancer great trade"* |
| Melatonin - dreams | 112 | No | *"i tried taking melatonin pills to help me sleep and i felt rested in the mornings but they gave me the weirdest dreams ever"* |
| Folic acid - autism | 100 | No | *"pregnant women who take too much folic acid increase chance of babies developing autism"* |
| Vitamin E – prostate cancer | 99 | No | *"inadequate intake of micro nutrients such as zinc selenium and vitamin e has been shown to predispose to prostate cancer"* |
| Niacin - flush | 94 | Yes | *"has anyone ever had a niacin flush i must of taken too much because oh boy am i itching and my face is ruby"* |
| Vitamin A - liver | 87 | No | *"vitamin a is toxic to liver in large doses and it is better to increase levels of beta carotene and let the body convert the required levels beta carotene itself"* |
| Vitamin D - falls | 58 | Yes | *"too much vitamin d may increase falls and fractures"* |
| Fish oil - burping | 50 | Yes | *"stupidly took fish oil tablets on an empty stomach and now burping through 2 hrs of meetings"* |

| DS Indication Pairs | Frequency | In iDISK? | Tweet examples |
|---|---|---|---|
| Melatonin - tired | 48 | No | "*melatonin has me feeling hella tired*" |
| Biotin - acne | 32 | No | "*started taking biotin to make my hair longer instead of long hair i have gotten acne and my leg hair grows at the speed of light*" |

Table 6. Examples of most frequently DS indication pairs detected by end-to-end deep learning pipeline.

| DS Indication Pairs | Frequency | In iDISK? | Tweet examples |
|---|---|---|---|
| Vitamin C - sick | 1924 | No | "*i am just going to eat 800 vitamin c today because i do not have time to be sick*" |
| Vitamin C - flu | 1043 | Yes | "*i would like to thank vitamin c for getting me through the flu season blessed*" |
| Vitamin C – colds | 759 | Yes | "*we ward off colds and flu with daily vitamin lots of liquids water and vitamin c rest and frequent hand washing s dm flu fighters*" |
| Biotin – hair growth | 653 | Yes | "*lol sick i just chopped mine off not too long ago go buy some biotin it promotes hair growth*" |
| Melatonin - sleep | 599 | No | "*i wanted to try to sleep without melatonin but i am realizing i can not so ill get on that*" |
| Vitamin D - pain | 328 | Yes | "*no wonder why i had all the weakness and pain in body now it makes sense vitamin d is three times below average and weak bones fml*" |
| Vitamin - tired | 322 | No | "*my body feels so tired weak i need me some vitamins preferably vitamin d*" |

# DISCUSSION

While many annotated corpora are readily available for drug AE extraction, they were not tailored for DS AE extraction. In this study, we demonstrated the feasibility of using Twitter as a complementary resource for DS AE surveillance. We thus created our own annotated tweet dataset to evaluate machine learning and deep learning models and develop an end-to-end DS AE signal detection pipeline. Our emphasis was to study the impact of traditional and contextual word embeddings on concept extraction and relation extraction performance. The evaluation results have shown that contextual word embeddings, i.e. BERT and BioBERT, outperformed traditional word embeddings in both concept extraction and relation extraction tasks regardless of the type of the classifier used in conjunction with the traditional word embeddings. This suggests that for concept extraction and relation extraction task, proper choice of word embeddings can lead to higher performance than using classifiers of complex architecture. Another advantage of BERT and BioBERT embeddings is versatility. For traditional embeddings, different tasks require different classifiers. For example, in this study, we used LSTM-CRF for concept extraction and CNN for relation extraction when using GloVe word vectors. However, when using BERT and BioBERT embeddings, we can keep the same deep neural network architecture and only fine-tune the model by training on different inputs.

We further compared the performance of BERT and BioBERT. As expected, BioBERT models outperformed BERT models in extracting DS and symptom entities. BioBERT models also outperformed BERT models in relation extraction tasks when no entity headings features were added. Especially, the BioBERT model was able to achieve the highest recall in supplement (0.8985) and symptom concept (0.8612) extraction. For body organ concepts, however, the BERT model outperformed the BioBERT model in precision, recall, and F1-score. One possible reason is that supplement and symptom terms appear more frequently in biomedical texts than common texts like Wikipedia. Body organ terms, on the other hand, are not rare in common texts.

We examined the DS AE relations retrieved from our proposed end-to-end pipeline and focused on corroborating the DS AEs that were not in iDISK. Two of the most frequently mentioned DS AEs in our dataset are 1) fish oil might cause prostate cancer and 2) melatonin might lead to nightmares. The relationship between fish oil usage and prostate cancer risk became popular among media based on the Brasky study, but there is no evidence that fish oil can cause prostate cancer. [42] Melatonin, on the other hand, is frequently used by people with sleeping disorders. [43] While it is still not certain whether the melatonin is the cause of nightmares and crazy dreams. These two examples have shown that our proposed model can detect the associations between DS and AE and it could be the basis of further clinical trials or safety test of the supplements.

**Error Analysis and Limitations**

Our end-to-end pipeline model was able to achieve almost equal mean F1-scores in identifying DS indications (0.7459) and DS AEs (0.7414), but the standard deviation of F1 score for identifying the DS AEs (0.0321) was almost double the one for DS indications (0.0699). While the decrease in F1-score was expected due to the propagation of concept extraction error through the pipeline, we studied the misclassification details of our pipeline from 10 out of 20 trained models.

The end-to-end pipeline model is prone to generate false positives than false negatives for DS indication identification. The pipeline has generated 89.2 false positive DS indications on average, which is more than 67.5 on average for false negative DS indications. This means the pipeline will be more likely to label a pair as DS indication when there is no relation at all. The false positives can be classified into two categories. 1) The pipeline correctly extracted the concepts but labeled a spurious relation. For example, in the tweet "'hi jeanette much better thanks upped my manuka and echinacea so just a bad cough now it will pass with time", the pipeline identified the DS "echinacea" and the symptom "cough" correctly, but no relation exists between these two concepts because neither did echinacea cause the cough nor treat the cough. However, the pipeline labeled a DS indication relation between these two. 2) The pipeline incorrectly extracted the concepts, resulting in a spurious relation. For example, in the tweet "'I have sun poisoning on my ass so there is that apple cider vinegar takes the pain away", the user used apple cider vinegar as a topical pain killer. However, the pipeline identified an extra symptom concept "sun poisoning" that did not exist in the gold standard, which results in a false positive for DS indication. These false positives contributed to a 10.5% decrease in the pipeline performance of DS indication relation extraction based on the F1-measure (from 0.8335 from standalone evaluation to 0.7459 from pipeline evaluation).

As for the DS AEs identification, the pipeline has generated 14.2 false positive DS AEs and 19.6 false negative DS AEs. The false negatives made greater impact on the performance. The false negatives could be classified into three categories. 1) The pipeline correctly extracted the concepts but labeled the relation wrong. For example, in the tweet "if you are experiencing diarrhea avoid greasy and fried foods caffeine sugary drinks and fruit juices healthy food", caffeine might lead to diarrhea and thus the relation between "caffeine" and "diarrhea" should be a DS AE. The pipeline identified the DS "caffeine" and the symptom "diarrhea" right but labeled the relation as DS indication; 2) The pipeline correctly extracted the concepts but did not give a relation label although a DS AE was in the gold standard. For example, in the tweet "kept on vomiting last night carbonated drinks and caffeine was on the do not drink list but I still wtf ok self stahp", the caffeine made the user vomit, therefore, the relation between "caffeine" and "vomit" should be DS AE, yet the pipeline did not give any label; 3) The pipeline failed to extract the concept that constitutes a DS AE relation. For example, in the tweet "I have not had crazy outbreaks from this biotin however I have noticed small acne flares very small", biotin could cause an outbreak, therefore, the relation between "biotin" and "outbreak" should be DS AE. However, the pipeline was not able to extract the entity "outbreak", which results in a false negative of DS AE. The misclassification of DS AEs only led to a 1.6% decrease in the pipeline F1-measure of pipeline DS AE identification.

The larger standard deviation in DS AE identification performance can be attributed to two reasons. 1) the dataset is highly imbalanced. We calculated the average ratio of DS indications and DS AEs among 20 test sets and the result was 3.31, meaning there were 3 times as many DS indications as DS AEs. The smaller sample size of DS AEs contributed to a higher standard deviation. 2) Among the 10 test sets we chose to analyze, two test sets yielded F1-scores of 0.6344 and 0.6255, respectively, for DS AE identifications. In both test sets, the pipeline was able to identify the concepts correctly (i.e. the pipeline was able to identify the supplement and symptoms) but failed to recognize negation of the concepts. For example, in the tweet "if ur not gettin' enough sun during the day, ur vitamin d level can drop and make you feel down and cause anxiety", Vitamin D is "deficient" and will lead to AEs "anxiety". The pipeline recognized Vitamin D as a supplement but did not extract the "deficiency" information. This would lead to an incorrect identification as the DS AE extracted from the pipeline suggests that "taking Vitamin D would lead to anxiety". Due to the scarcity of tweets that contain deficiencies in the training set, a poor randomization could split most of the data points that contain negations into the test set instead of the training set, which results in inferior performance in identifying relations that contains information about DS deficiencies.

## CONCLUSION

We developed an end-to-end deep learning pipeline to identify AE of DS from tweets. We compared the effect of word embeddings and classification methods on the model performance for each task to find the best candidates for the end-to-end DS AEs identification pipeline. We assembled the end-to-end pipeline with these two best models and the model could identify DS purposes and AEs with F1-score of 0.8130 and 0.7065, respectively. The trained end-to-end pipeline was applied to the entire dataset and the DS AEs extracted by the pipeline was compared to iDISK. We found that our proposed machine learning pipeline not only retrieved DS AEs recorded in the current DS knowledge database, but also discovered DS AEs that were only reported in tweets. The result suggests that Twitter is indeed a complementary source for monitoring DS AEs and our pipeline can detect these signals for further clinical trials or safety research.

## ACKNOWLEDGEMENTS

We would like to thank Dr. Rubina Rizvi for her annotation of the tweets.

## FUNDING STATEMENT

This work was supported by the National Institutes of Health's National Center for Complementary & Integrative Health (NCCIH) and the Office of Dietary Supplements (ODS) grant number R01AT009457 (Zhang),

**COMPETING INTERESTS STATEMENT**

The authors state that they have no competing interests to declare.

**CONTRIBUTORSHIP STATEMENT**

RZ and JB conceived the study design. YW carried out the experiments and produced the original draft of the manuscript. All authors contributed to the production of the manuscript.